\documentclass{IEEEoj}
\usepackage{cite}
\usepackage{amsmath,amssymb,amsfonts}
\usepackage[ruled,vlined]{algorithm2e}
\usepackage{graphicx,color}
\usepackage{multirow} 
\usepackage{textcomp}
\renewcommand{\OJlogo}{}

\begin{document}
\receiveddate{XX Month, XXXX}
 \reviseddate{XX Month, XXXX}
% \accepteddate{XX Month, XXXX}
% \publisheddate{XX Month, XXXX}
% \currentdate{XX Month, XXXX}
%\doiinfo{OJITS.2022.1234567}

\title{Uncertainty-Aware Adaptive Sensor Fusion for Autonomous Navigation}

 \author{Simegnew Yihunie Alaba, \IEEEmembership{Senior Member, IEEE} and Yuichi Motai, \IEEEmembership{Senior Member, IEEE}
 \affil{Department of Electrical and Computer Engineering, College of Engineering, Virginia Commonwealth University, Richmond, VA 23220, USA}}
% \corresp{CORRESPONDING AUTHOR: Yuichi Motai (e-mail: ymotai@vcu.edu).} }
% \authornote{This work was supported in part through Minority Serving Program funded by the Department of Defense under Grant W911NF231016.}
\markboth{Uncertainty-Aware Adaptive Sensor Fusion for Autonomous Navigation}{Alaba \textit{et al.}}

\begin{abstract}
This work introduces a hybrid deep learning approach integrated with an Unscented Kalman Filter (UKF) to enhance pose estimation accuracy in Visual-Inertial Odometry (VIO) for autonomous navigation. The proposed model employs a Vision Transformer (ViT) network to effectively capture temporal dependencies from inertial measurement unit (IMU) data and utilizes a Multiscale Convolutional Neural Network (MCNN) to learn optical flow-based motion cues from visual data. An adaptive sensor fusion module dynamically weights IMU and visual features by leveraging estimated uncertainty, thus improving robustness in diverse and challenging environmental conditions. Additionally, a novel uncertainty-aware loss function is proposed to explicitly incorporate prediction uncertainty into the learning process, enabling robust and accurate navigation under noisy, incomplete, or unreliable sensor inputs. Comprehensive evaluations of the KITTI dataset demonstrate that the proposed method significantly outperforms baseline approaches, achieving superior performance in terms of Absolute Trajectory Error (ATE) and Relative Pose Error (RPE). The lightweight and computationally efficient model processes data at 155 FPS on an NVIDIA A100 GPU, making it highly suitable for deployment in resource-constrained autonomous systems. The code will be released.
\end{abstract}



\maketitle

%%%%%%%%%%%%%%%%%%%%%%%%%%%%%%%%%%%

\section{Introduction}
\IEEEPARstart{I}{n } spite of significant progress in sensor technology and machine learning, achieving reliable and precise localization in complex environments remains challenging for autonomous vehicle and robotics navigation. Inertial Measurement Units (IMUs) are crucial in motion tracking by utilizing acceleration and angular velocity data. However, the effectiveness of IMUs is limited by their susceptibility to drift over time, which accumulates errors in velocity and position estimations from acceleration data \cite{qin2018vins}. On the other hand, visual sensors provide rich environmental data by capturing detailed information about the vehicle’s surroundings. However, visual sensors are affected by environmental conditions, such as occlusions, lighting variations, and limited field of view, and cannot provide reliable pose estimates when visual features are scarce or distorted. As a result, sensors come with limitations, and no single sensor is suitable for every application \cite{alaba2023deep}.

To address these challenges, sensor fusion techniques have shown promising results. Visual-inertial odometry (VIO) is one common approach, fusing visual data with inertial measurement to enhance localization. Cameras provide detailed scene information, improving position tracking accuracy by compensating for IMU measurement drift. While traditional VIO methods, which depend on handcrafted features and fuse features using  Bayesian filters, \cite{qin2018vins, forster2016manifold, leutenegger2015keyframe, li2013high, wan2000unscented} have shown potential, they often require manual calibration and struggle in challenging conditions such as low light or fast movement \cite{yang2018challenges, el2007analysis}. Recent advancements in deep learning (DL), particularly convolutional neural networks (CNNs), have greatly enhanced feature extraction, offering more robust solutions for VIO \cite{chen2019selective, yusefi2023generalizable, clark2017vinet, kendall2015posenet, clark2017vidloc}. However, these models may still encounter difficulties when generalizing to diverse environments.

This work presents an adaptive sensor fusion model that integrates DL with the Unscented Kalman Filter (UKF) to fuse IMU and visual data for autonomous vehicle navigation. The DL model learn features and predicts pose, such as position, velocity, and orientation, from the sensors data, and the UKF uses the pose estimates to produce a final localization estimate and refine it.
The model is designed to be lightweight, incorporate uncertainty estimation, and dynamically adjust sensor noise to handle dynamic environmental changes. This adaptability is crucial as IMU and visual sensor data can vary in reliability depending on the conditions. Uncertainty estimation measures the confidence in the system's predictions, helping to quantify how much trust should be placed in the predicted pose estimation. This adaptability enhances the robustness of the navigation system, particularly in dynamic or complex environments.
Additionally, an uncertainty-aware loss function that incorporates uncertainty into the overall loss calculation is proposed. This ensures that the system considers potential errors in sensor data during the learning process and refining its predictions based on data reliability. This work also introduces feature extraction networks to learn the essential features from sensors. A Vision Transformer (ViT) network is proposed to process and learn important features from IMU data, capturing temporal dependencies more effectively than traditional methods such as Long Short-Term Memory (LSTM) networks due to its ability to handle long-range dependencies, which handles global information and reduces accumulated errors, and multihead attention mechanism for faster processing speed \cite{vaswani2017attention}. Similarly, a multiscale convolutional neural network (MCNN) has been introduced to extract essential features from visual data before feeding them into the UKF. This model is designed to learn features at various scales using a multiscale module and addresses the issue of vanishing gradients with a residual block.

This adaptive fusion method enables the UKF to estimate noise covariance adaptively by capturing complex patterns in the data based on the current system state and sensor characteristics, eliminating the need for static or manual tuning based on prior system knowledge. The proposed work, validated on the KITTI visual-odometry dataset \cite{Geiger2013IJRR}, significantly reduces position estimation errors. 
This error reduction demonstrates the adaptive fusion approach's effectiveness in offering a reliable solution for autonomous vehicle navigation in complex environments.
 
The main contributions of our work are summarized as follows:
\begin{enumerate} 
\item We present a DL-based approach for feature learning and fusion to improve navigation accuracy and reliability in autonomous driving systems.
\item We incorporate aleatoric and epistemic uncertainties into the pose estimation process to enhance reliability. By introducing an uncertainty-aware loss function, we dynamically adjust parameters based on uncertainty levels, allowing for more accurate and robust predictions, even when sensor data is weak or inconsistent.
\item We develop an adaptive sensor fusion method with a gate-based feature scaling mechanism. This allows us to dynamically weigh the importance of different features, emphasizing the most relevant ones and improving the fusion process. 
\item We combine the DL approach with UKF to achieve more accurate pose estimation and refine the overall process for better autonomous vehicle navigation.
\item We developed a ViT network to process IMU data, which enables it to capture temporal dependencies more effectively than traditional methods such as  LSTM networks. Similarly, a multiscale CNN network is developed to learn visual features. 
\end{enumerate}

The remaining sections of the paper are structured as follows: In Section \ref{sec:related}, the related work on autonomous driving navigation is presented. Section \ref{sec:method} provides an in-depth explanation of the proposed model's architecture, including the DL feature extractor, uncertainty estimation, adaptive fusion, Unscented Kalman filter, and uncertainty-aware loss function used for model training. Section \ref{sec:experiments} showcases the experimental results and analyses, along with details of the training setup. Finally, Section \ref{sec:conclusion} summarizes the experimental results.

\section{Literature Review}
\label{sec:related}
This section presents various multisensor fusion-based navigation methods, including traditional (handcrafted), DL-based, and Bayesian filter-based navigation.
\subsection{Handcrafted Feature-based Navigation}
These traditional methods rely on handcrafted feature extractors, such as Scale-Invariant Feature Transform (SIFT) \cite{lowe2004distinctive}, Speeded Up Robust Features (SURF) \cite{bay2006surf}, and Oriented FAST and Rotated BRIEF (ORB) \cite{rublee2011orb}  to find keypoints. After detecting features in consecutive frames, matching is performed using algorithms, such as  Kanade–Lucas–Tomasi (KLT) \cite{tomasi1991detection}, Farneback Optical Flow \cite{farneback2003two} to establish correspondences between the same features in different frames. This matching helps track how points in the environment move relative to the camera or vehicle.
SVO \cite{forster2014svo} was introduced to calculate pose using a semi-direct method that combines the direct photometric error minimization with the handcrafted feature-based techniques. SVO offers a solution for visual odometry by tracking key points and improving depth and pose through triangulation and bundle adjustment. However, it has limitations due to monocular scale ambiguity and sensitivity to changes in lighting.
Similarly, Younes et al. \cite{younes2018fdmo} presented a hybrid approach that combines monocular visual odometry and feature-based methods. By leveraging pixel intensities and keypoint information, the proposed method provides a more reliable solution for pose estimation in monocular systems. However, like other monocular systems, it suffers from scale ambiguity and can still be affected by poor feature detection in certain scenarios.
Silva et~al. \cite{silva2015probabilistic} proposed a probabilistic ego-motion model for stereo-visual odometry to improve pose estimation. Stereo vision provides precise depth information, which makes pose estimation more reliable than monocular systems. However, the improved robustness of the probabilistic model comes with the trade-off of increased computational cost.

The main drawback of handcrafted-based visual odometry for pose estimation is its susceptibility to environmental factors such as low-texture scenes, changes in lighting, and rapid motion. These methods heavily depend on manually crafted features like edges, corners, or keypoints, which are identified using SIFT, SURF, or ORB algorithms. These methods struggle to locate reliable tracking points in environments with few distinctive features, such as low-texture surfaces. Additionally, sudden changes in lighting, fast camera movement, or obstructions can result in feature mismatches or tracking failures.
Moreover, handcrafted-based methods typically involve a sequence of feature detection, matching, and geometric optimization, which requires manual adjustments and can be computationally expensive. They are also less robust to variations in scale and rotation, especially when the conditions deviate from those for which they were explicitly designed, leading to reduced performance in dynamic or unpredictable environments.

\subsection{Deep Learning-based Navigation}
In DL-based methods, models automatically learn features without requiring manual feature extraction.  For instance, Kendall et al. \cite{kendall2015posenet} introduced PoseNet, a DL model designed to estimate the camera pose from a single RGB image. PoseNet utilizes CNNs to compute the camera's position and orientation. Similarly, HVIOnet \cite{aslan2022hvionet} fuses visual and inertial data using a neural network to estimate the position of unmanned aerial systems.
 Wu et al. \cite{wu2017delving} proposed CNN-based model that enhance camera pose estimation by capturing richer spatial and contextual information. VidLoc  \cite{clark2017vidloc} utilizes spatial and temporal data from video sequences to estimate camera pose.  Zhou et al. \cite{zhou2017unsupervised} proposed an unsupervised deep learning approach capable of simultaneously estimating depth and ego-motion from monocular video sequences, eliminating reliance on ground-truth data and improving generalization to unseen environments.  Similarly, Zhan et al.  \cite{zhan2018unsupervised} presented an unsupervised method that integrates deep feature reconstruction into monocular depth estimation and visual odometry, significantly enhancing depth prediction accuracy and odometry robustness without labeled data.
Moreover, Brahmbhatt et al. \cite{brahmbhatt2018geometry} combined DL with geometric reasoning to enhance localization. The method involves learning geometry-aware maps incorporating visual and geometric information from the environment to improve relocalization performance.
Additionally, Valada et al. \cite{valada2018deep} presented a multitask learning strategy to improve visual localization and odometry. Incorporating auxiliary tasks such as scene classification and depth prediction alongside the primary task of camera pose estimation enables the network to learn more robust and transferable features. Xue~et al.\cite{xue2020deep} introduced a recurrent neural network-based visual odometry that comprises adaptive memory for remembering and refining.  Chen et al. \cite{chen2019selective} presented an end-to-end VIO fusion using an LSTM network. Gao et al. \cite{gao2023efficient} proposed a hierarchical reinforcement learning model for mapless robot navigation, utilizing predictive neighboring space scoring to enhance navigation efficiency and robustness. This method exploits a deep reinforcement learning paradigm for robust collision avoidance and improved path planning without relying on prior environmental mapping. Recently, Nayak et al. \cite{nayak2022uncertainty} estimate predictive uncertainty in pedestrian trajectory forecasting using Bayesian  deep learning approximations such as Monte Carlo dropout, allowing autonomous vehicles to plan more robust trajectories in dynamic scenes. Generative multi-modal fusion has been explored for better detection in autonomous driving, where a generative sensor-fusion model is used to identify out-of-distribution situations from multi-sensor inputs. Ni et al. \cite{ni2024adaptive} use deep reinforcement learning to design an adaptive cruise control system that exploits noisy distance and speed measurements from both first and second leaders, improving string stability and ride comfort under measurement uncertainty.

\subsection{Bayesian Filter-based Navigation}
Most fusion-based localization works use Bayesian filters for autonomous vehicles and robotics navigation. For example, the fusion of Global Navigation and Satellite System (GNSS) and IMU addresses limitations when these sensors operate independently, particularly in environments with weak or blocked GPS signals, such as urban areas or indoor settings. 
Various filtering techniques are used to fuse GNSS/GPS and IMU data effectively, with Kalman Filters (KF) \cite{kalman1960contributions} and their variants, such as the Extended Kalman Filter (EKF), the Unscented Kalman Filter (UKF), etc. Caron {\em et al.} \cite{caron2006gps} introduced a multisensor KF technique incorporating contextual variables to improve GPS/IMU fusion reliability, especially in signal-distorted environments. Lee {\em et al.} \cite{lee2016camera} put forth a sensor fusion method that combines camera, GPS, and IMU data, utilizing an EKF to improve state estimation in GPS-denied scenarios. Similarly, Suwandi {\em et al.} \cite{suwandi2017low} demonstrated a cost-effective approach to vehicle navigation by focusing on low-cost IMU and GPS sensor fusion to improve navigation. Atia {\em et al.} \cite{atia2017low} combined MEMS, IMU, GPS, and road network maps with an EKF and Hidden Markov model-based map-matching to provide accurate lane determination without high-precision GNSS technologies. Li and Xu \cite{li2016reliable} introduced a method for sensor fusion navigation in GPS-denied areas. This method fuses sensors with a sliding mode observer and a federated KF. Yáng and Chen \cite{yang2023navigation} proposed a probabilistic navigation method in uncertain environments through overlap-induced equilibrium points. This method dynamically optimizes navigation paths and ensures robust collision avoidance, effectively addressing challenges in coordinated multi-robot navigation under environmental uncertainty and variability.

Liu {\em et al.} \cite{liu2018innovative} developed an enhanced adaptive KF with an attenuation factor to handle noise effectively to improve navigation accuracy. Meng {\em et al.} \cite{meng2017robust} employed the GNSS, IMU, DMI, and LiDAR to counteract inaccuracies caused by GNSS signal jumps and multipath interference in urban settings. Tao {\em et al.} \cite{tao2021multi} introduced a multisensor fusion strategy that fuses GNSS, IMU, and visual data with global pose graph optimization using the KITTI dataset. Yusef {\em et al.} \cite{yusefi2023generalizable} employed a deep VIO-based model to improve accuracy in low-GNSS areas. Similarly, Park \cite{park2024optimal} used an adaptive Kalman filter for vehicle position estimation to address GPS outages. Gruyer and Pollard \cite{gruyer2011credibilistic} enhanced navigation in GPS-denied environments using proprioceptive sensors with an Interacting Multiple Model (IMM) filter. Some works \cite{bai2021improved, qin2018vins} proposed graph-based optimization for autonomous vehicle navigation. Qiu et al. \cite{qiu2024outlier}  developed an outlier-robust EKF  for bioinspired integrated navigation systems. The method integrates polarization sensors to effectively handle measurement outliers, significantly improving navigation accuracy and reliability. Wang et al.  \cite{wang2025relative} addressed the challenge of navigation under non-persistent excitation conditions by integrating Ultra-Wideband (UWB) and IMU measurements. 
Similarly, Xiong et al. \cite{xiong2025gfslam} introduced GF-SLAM, a hybrid navigation incorporating global positioning and arc-shaped environmental features into a feature-based SLAM using an EKF. Geiger et al. \cite{geiger2011stereoscan} introduced Stereoscan, a real-time system for dense 3D reconstruction using stereo imagery, enhancing spatial perception in intelligent vehicles. Mur-Artal and Tardós \cite{mur2017orb} developed ORB-SLAM2, an open-source SLAM system supporting monocular, stereo, and RGB-D cameras, providing robust performance for real-time localization and mapping tasks. Recently,  Harbers et~al. \cite{harbers2025vehicle} present an IMU–RADAR fusion with Kalman-filter-based state estimation for robust motion estimation in challenging conditions. Our method is complementary, as we fuse IMU and camera features and additionally incorporate aleatoric and epistemic uncertainty into an adaptive UKF‑based fusion framework with an uncertainty‑aware loss function. Iqbal et al. \cite{iqbal2025novelty} propose a generative multi‑modal sensor fusion framework for autonomous driving that fuses proprioceptive (wheel‑odometry) and exteroceptive (LiDAR) data and applies Bayesian filtering to detect novel situations under environmental uncertainty.

Although these studies propose novel methods to enhance navigation, the reliability of navigation results remains a challenge, particularly in cases of sensor failure or when sensor data is weak or noisy. This limitation can lead to inaccuracies in pose estimation, which is critical for safe and effective navigation. To address this issue, we introduce uncertainty quantification, incorporating both aleatoric and epistemic uncertainties into the navigation model. 
By quantifying both uncertainties, we can better understand the confidence in the model's predictions and make more informed decisions about pose estimation.
Furthermore, we integrate this uncertainty quantification into the model's training process through an uncertainty-aware loss function. This allows the model to dynamically adjust its parameters based on uncertainty, leading to more accurate and reliable predictions, even when sensor information is weak or inconsistent. The uncertainty-aware loss function prioritizes minimizing uncertainty in predictions, thus improving the model's robustness under challenging conditions.
Integrating uncertainty quantification is particularly important for complex systems where sensor information may be noisy, unreliable, or temporarily unavailable. Explicitly incorporating uncertainty enhances the model's adaptability and provides more reliable navigation performance even in adverse conditions. This approach not only improves detection but also helps determine how much trust can be placed in the pose estimation, leading to a more robust and adaptive autonomous navigation system that can be relied upon in challenging conditions.
 
Moreover, the proposed adaptive fusion method introduces a unique gate-based feature scaling mechanism with concrete distribution. This mechanism, unlike traditional methods that treat all inputs equally, dynamically weighs the relevance of various features. By utilizing gates to emphasize the most significant and relevant features, the system enhances those that contribute most to accurate pose estimation, providing a more refined method for feature selection and fusion.
This approach improves the feature fusion process by selectively filtering out less valuable or noisy inputs while focusing on high-quality data. This is particularly crucial when dealing with high-dimensional inputs from multiple sensors, where a large amount of data can be challenging to manage and prone to information redundancy or noise. By applying this feature scaling, the system gains more control over the fusion of these complex datasets, allowing it to prioritize critical information for decision-making intelligently. This DL-enhanced pose estimation technique improves the accuracy and reliability of autonomous vehicle navigation.

\section{Method}
\label{sec:method}
\begin{figure}[!ht]
    \centering
    \includegraphics[width=0.99\linewidth]{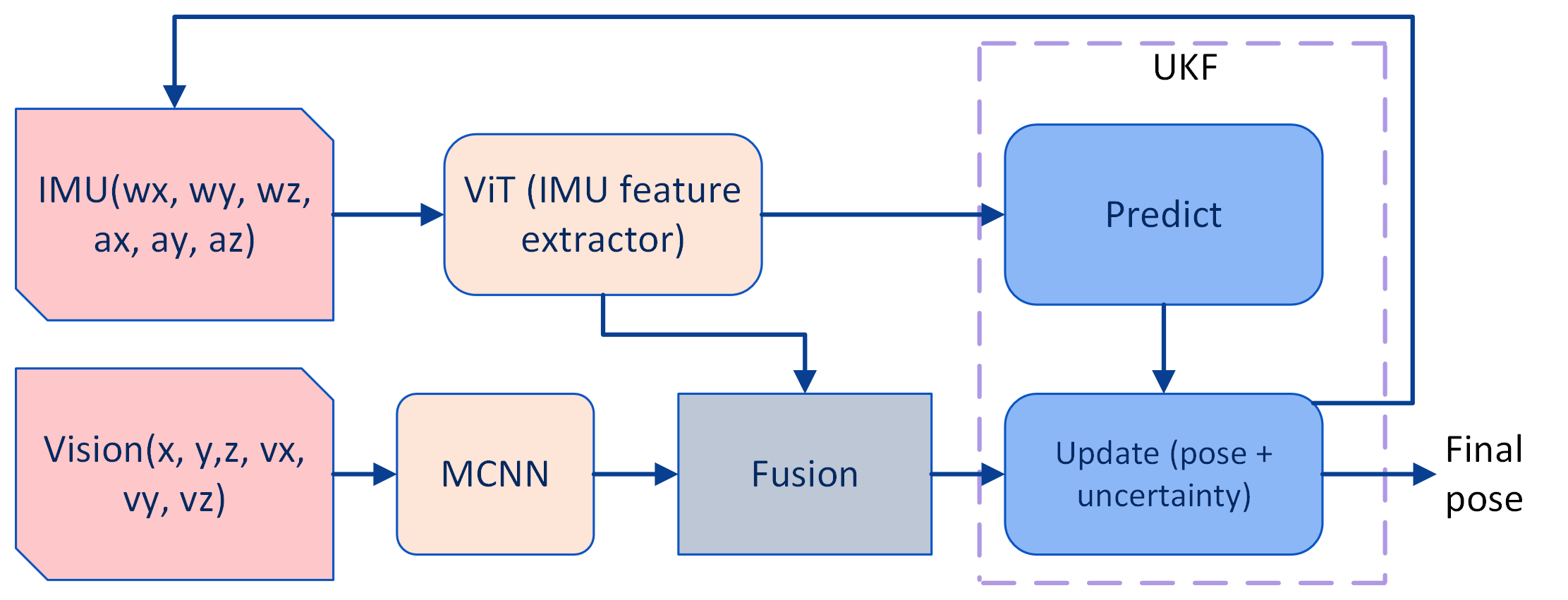}
    \caption{The proposed architecture. The ViT learns essential features from the IMU, while the MCNN learns important visual features and motion-related information from visual odometry. The UKF predicts the pose, refines it, and corrects the errors.  Where $\omega_x$, $\omega_y$, and $\omega_z$ represent rotational movement, and $a_x$, $a_y$, and $a_z$ denote acceleration.
}
    \label{fig:proposed}
\end{figure}

This section presents VIO-based fusion for autonomous driving navigation. The proposed architecture includes a  MCNN for visual feature extraction, a ViT for IMU feature extraction, an adaptive fusion network, and the UKF's prediction and update steps., as shown in Fig \ref{fig:proposed}. 
The proposed work presents a sensor fusion model for autonomous vehicle navigation, integrating data from two primary sources: an IMU and visual sensors (camera). This fusion of IMU and visual data leverages the strengths of each sensor while compensating for their respective limitations, ensuring more robust and accurate vehicle navigation, even in challenging and dynamic environments.
With its accelerometers and gyroscopes, the IMU provides real-time information on the vehicle's acceleration and rotational movements, offering continuous data crucial for maintaining orientation and estimating movement even when external references are unavailable. However, the IMU suffers from drift over time, accumulating position and orientation estimate errors. Without correction, this drift can reduce the accuracy of the navigation system, especially over long distances.

On the other hand, visual sensors provide rich environmental data, capturing detailed information about the vehicle's surroundings. Cameras can identify landmarks, track features, and estimate motion through techniques like visual odometry. However, visual sensors are susceptible to environmental conditions, such as occlusions, lighting variations, and limited field of view, and cannot provide reliable pose estimates when visual features are scarce or distorted but can provide absolute references to correct the drift experienced by the IMU.
An adaptive fusion method is proposed to fuse the high-rate IMU and visual data to handle these limitations and make the most out of the two sensors. Before fusing, DL techniques are used to extract features that significantly enhance pose estimation accuracy. Specifically, a ViT network is introduced to capture crucial features from the IMU data, identifying important temporal patterns that contribute to better motion estimation. Meanwhile, a  MCNN, inspired by FlowNet \cite{dosovitskiy2015flownet}, is introduced to process visual information, learning optical flow to track the relative motion of the vehicle within its environment.

DL techniques are essential for learning complex features that might not be directly observable from raw IMU and visual data. The ViT enhances the IMU data by extracting high-level representations, making it more reliable for pose estimation. Additionally, the FlowNet-based MCNN effectively captures motion cues from the visual data, complementing the IMU data, particularly when the IMU experiences drift.
The learned IMU features are first used to estimate the vehicle's pose, which provides an initial prediction of its position and orientation. The fusion of IMU and visual features is then used in the UKF update phase, refining the pose estimate. This fusion provides a balanced approach—leveraging the continuous, high-rate IMU data for pose estimation and the detailed, external references from visual data to correct errors and ensure accuracy. Additionally, uncertainty estimation is explicitly included during sensor fusion and update stages. Estimating uncertainty provides confidence in the system's predictions and determines the level of trust to place in the predicted pose estimation. This adaptability enhances the navigation system's robustness. An uncertainty-aware loss function is proposed to ensure that the system considers potential errors in sensor data during the learning process and refines its predictions based on data reliability.

This approach showcases a combination of conventional filtering techniques (UKF), uncertainty estimation, and DL, providing a reliable solution for autonomous navigation, especially when a single sensor is insufficient. The proposed model ensures the vehicle can safely navigate complex environments with dynamic updates to its pose, improving decision-making and enhancing overall navigation safety.

\subsection{Multiscale Deep Learning Network for Visual Feature Extraction }
\begin{figure}[!ht]
    \centering
    \includegraphics[width=0.98\linewidth]{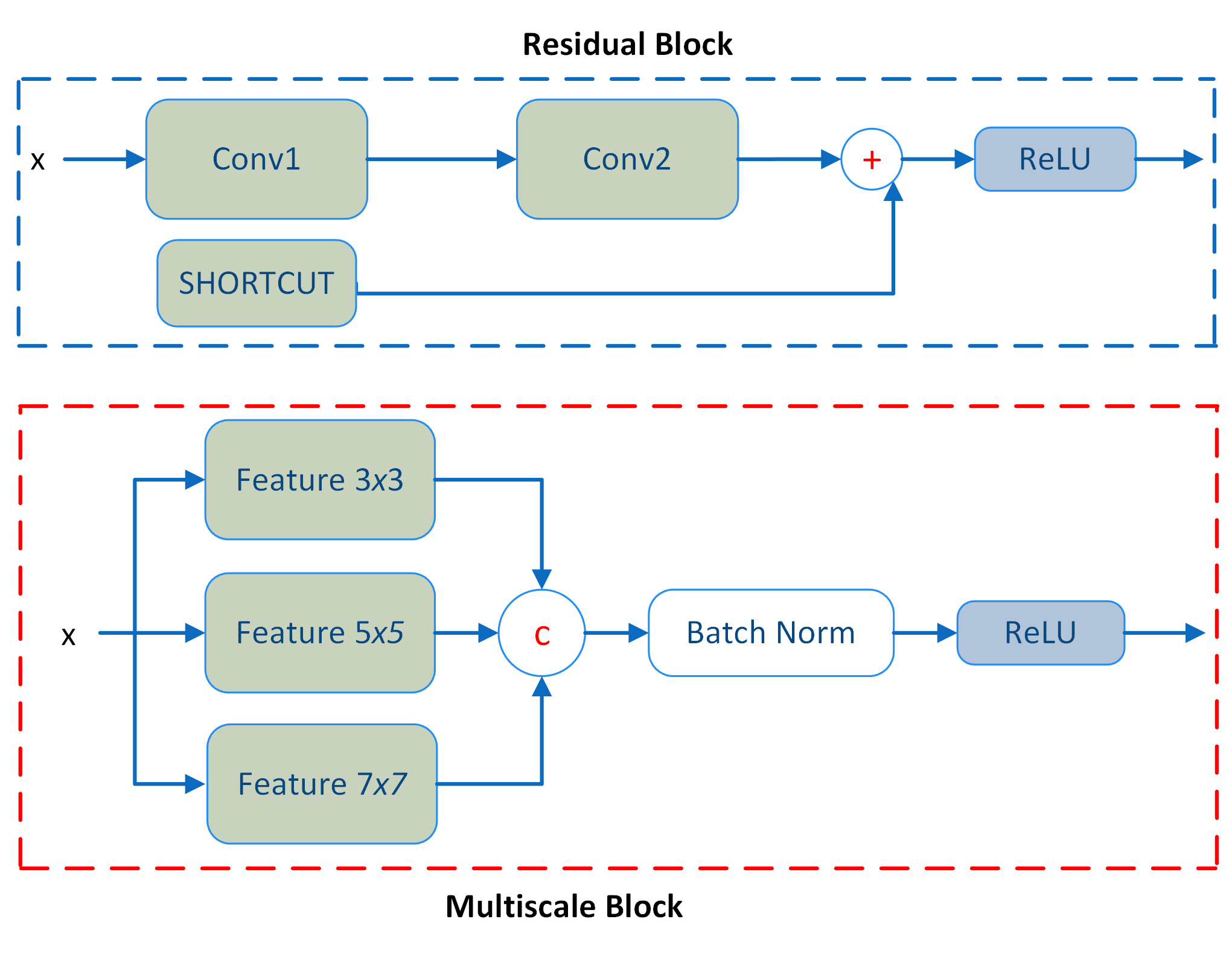}
    \caption{The proposed residual and multiscale blocks. Where {\em x} is an input to the block and  {\em \textcolor{red}{c}} and \textcolor{red}{+} represent concatenation and element-wise summation, respectively.}
    \label{fig:multiscale}
\end{figure}

The proposed MCNN is based on FlowNet \cite{dosovitskiy2015flownet}. The FlowNet network is designed to learn geometric features that are well-suited for predicting optical flow. 
The network comprises nine convolutional layers, with the receptive field sizes gradually decreasing from 7 $\times$ 7 to 5 $\times$ 5 and 3 $\times $ 3, using a stride of two for the first six layers. To address the computational expense of these large kernel sizes, the residual and multiscale blocks are introduced to reduce the computational load while extracting multiscale geometrical features for optical flow prediction, as shown in Fig. \ref{fig:multiscale}.
The residual block consists of two convolutional layers, each with a 3×3 kernel, followed by a Rectified Linear Unit (ReLU). A residual connection is added to enhance feature representation and address the vanishing gradient problem. 
The multiscale block comprises feature sizes of 7 $\times $ 7, 5 $ \times $ 5, and 3 $\times $ 3. The standard convolutions were replaced with depthwise separable convolutions to reduce the computational cost \cite{simonyan2014very}, splitting the convolution into two stages: a depthwise convolution and a pointwise (1 $\times $ 1) convolution. This approach significantly reduces the number of computations while maintaining accuracy. This multiscale feature extractor network learns latent features from a set of two consecutive images. By incorporating these two new blocks, the new network enhances detection with reduced computational load.

\subsection{Vision Transformer Network for Inertial Feature Extraction}
Inertial data comprises measurements from the accelerometer and gyroscope with a temporal component recorded at a higher frequency of 100Hz than the 10Hz frequency of image data. Inspired by the success of transformer networks in natural language processing and computer vision \cite{vaswani2017attention, dosovitskiy2020vit, alaba2024transformer}, we introduced a ViT network to extract important features from IMU sensor data. ViTs have demonstrated superior performance to LSTM in processing time series and image data, due to their ability to manage long-range dependencies effectively \cite{vaswani2017attention}. The multihead attention mechanism also allows for parallel processing, making computation significantly faster.  Each attention head has its own set of learnable parameters, enabling it to capture distinct patterns within the input sequence. The scaled dot-product attention mechanism involves queries (Q), keys (K), and values (V) as input, where both the queries and keys have dimensions of $d_{k}$ and the values have dimensions of $d_{v}$. The attention for each head can be computed  as follows:
\begin{equation}
  Attention(Q, K, V) =softmax(\frac{Q_{i} K_{i}^{T}} {\sqrt{d_{k}}})V_{i},
  \label{eqn:concat}
\end{equation}
Then, each head's attention output is concatenated. 

\subsection{Adaptive Fusion Network}
\begin{figure}[!ht]
    \centering
    \includegraphics[width=0.96\linewidth]{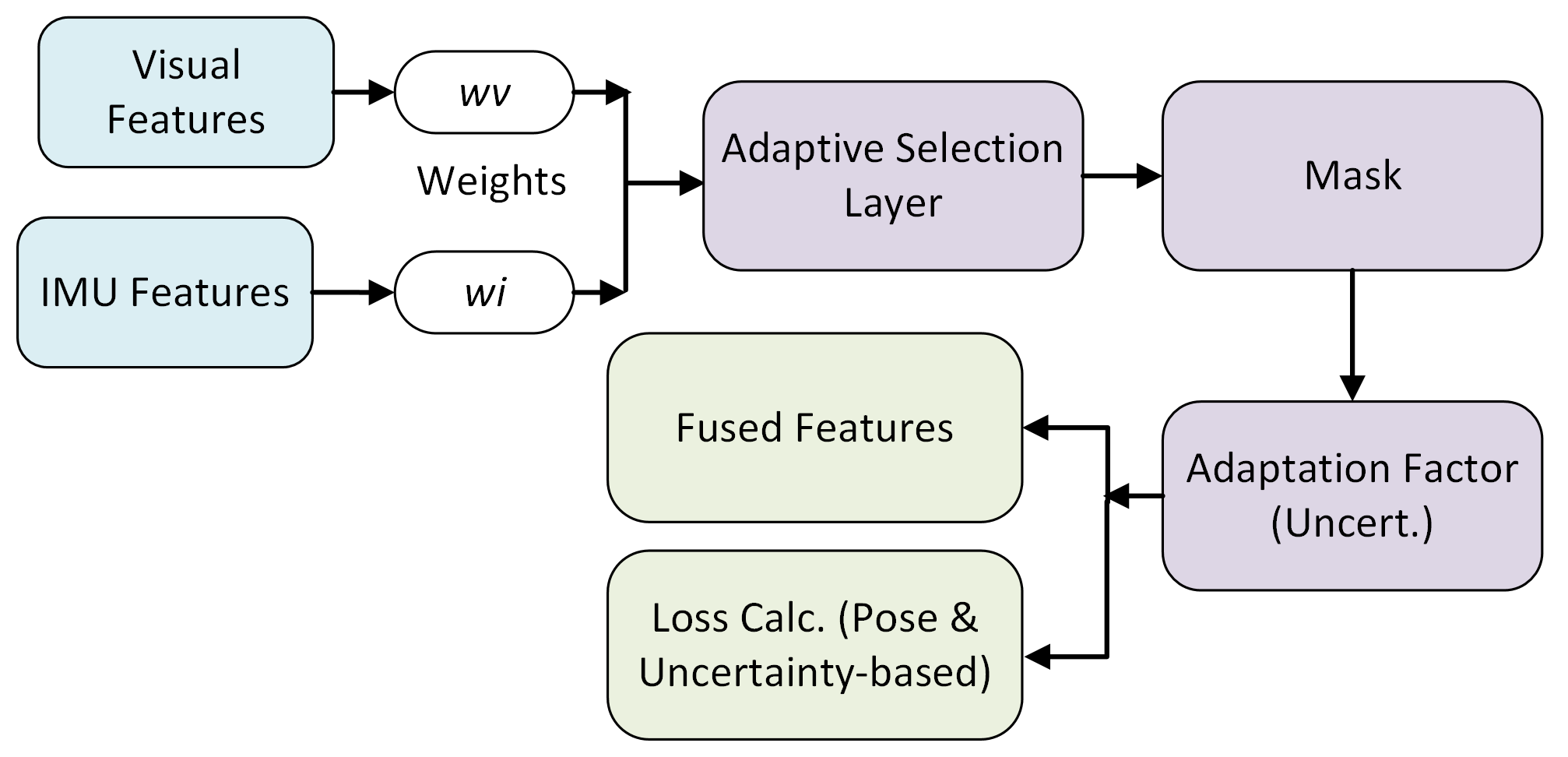}
    \caption{Adaptive fusion. The adaptive selection adaptively selects features based on their importance, followed by the adaptation factor adjustment. }
    \label{fig:fusion}
\end{figure}
As shown in Fig. \ref{fig:fusion}, the proposed uncertainty-aware adaptive fusion module integrates features from VIO using an adaptive mechanism that leverages uncertainty estimation to enhance robustness. Fusing sensor features with equal weights often leads to suboptimal results, mainly when one sensor provides limited, corrupted, or unavailable data. Therefore, assigning weights to features based on their contribution to pose estimation is more effective. This module introduces an adaptive weighting mechanism to dynamically adjust the contribution of IMU and visual features during fusion. A mask selectively emphasizes relevant features from both sensors, with weights generated based on the current features being processed. This approach enables the fusion process to adapt to varying conditions, such as sensor failure or data corruption. The mask is guided by uncertainty estimation to ensure that decisions are based on confidence in each sensor's data. For instance, if one sensor exhibits high uncertainty, the mask reduces its influence, giving greater weight to the more reliable sensor. The mask is generated using the Concrete distribution \cite{maddison2016concrete}, enabling learning a stochastic function to emphasize or downplay features. The Concrete distribution allows differentiable backpropagation by offering a continuous approximation of discrete latent variables. This approach uses a temperature parameter with temperature annealing, enabling a controlled, smooth transition from soft to hard sampling during training. While other techniques, such as the Straight-Through Estimator \cite{bengio2013estimating} and Stochastic Policy Gradient \cite{mnih2014neural}, can be applied, the Concrete distribution is lightweight, has low gradient variance, and facilitates smoother training, making it the preferred choice in this work.

The final fused features can be represented as follows:
\begin{equation}
    f_{fused} = w * f_{IMu} + (1- w) * f_{visual}
\end{equation}
Where {\em w} represents the adaptive weight used to balance the contribution of each modality. 

The adaptation factor is driven by uncertainty estimation to control the mask and determine the appropriate weight for each sensor. This adaptive fusion mechanism selects the most relevant features from each sensor, improving model performance by emphasizing essential features and disregarding irrelevant ones.
By dynamically adjusting the contribution of each sensor's data through the adaptive factor and incorporating uncertainty estimation to account for noise, the model ensures a more reliable and accurate fusion of features. Additionally, a new loss function incorporating uncertainty for the overall model is proposed to enhance feature learning and fusion further, improving the model's ability to handle uncertain or noisy sensor data for more accurate and robust fusion performance. The detail is presented in section \ref{uncertainty}.

\subsection{Unscented Kalman Filter}
The UKF is a widely used algorithm for estimating the state of a nonlinear dynamic system. The UKF extends the traditional Kalman Filter by better-handling nonlinearity in both the process and measurement models, offering a more accurate approach for state estimation. Unlike the extended Kalman filter (EKF), which relies on linear approximations, the UKF leverages a deterministic sampling technique to effectively capture the system states' mean and covariance. This makes it particularly useful for systems with significant nonlinearity, such as those encountered in autonomous navigation and robotic control.

The mathematical representation of the system comprises process and measurement models. The dynamic connection between the states at two consecutive time steps is managed by the process truth model, which can be defined as follows:

\begin{equation}
x_t = f\left( x_{t-1}, u_{t-1} \right) + w_{t-1},
\end{equation}
where \(x_{t}\) indicates the estimated state after an interval \(\sigma\), derived from the preceding state vector \(x_{t-1}\). The variable \(u_{t-1}\) functions as the input to the state space equations, and \(w_{t-1}\) reflects the noise affecting the process.

The measurement model for the visual data defines the relationship between the state vector and sensor measurements as follows:
\begin{equation}
    y_v = v + n_v,
\end{equation}
where \(v\) represents the visual data, and \(n_v\) denotes the noise associated with the visual measurements. The proposed MCNN extracts visual features and captures optical flow between consecutive images. These visual features and learned IMU features are fused adaptively using DL techniques, and the updated features are fed to the UKF for improved pose estimation.

The UKF's strength lies in its efficient use of the unscented transform. This involves selecting a minimal set of sample points around the mean, accurately capturing the state distribution's mean and covariance. These points are then propagated through the non-linear system, preserving the distribution's properties more effectively than linearization and enhancing the UKF's robustness.

\textbf{Prediction step:} The prediction step involves calculating sigma points and updating the process by predicting mean and covariance \cite{noureldin2012fundamentals}. The sigma points are calculated using:
\begin{equation}
\bar{X}_{t-1} = \begin{bmatrix} x_{t-1} & x_{t-1} \pm \sqrt{(n + \kappa) P_{t-1}} \end{bmatrix},
\end{equation}
where $\bar{x}_{t-1}$ denotes the sigma points of the state vector $\mathbf{x}$ at the prior time step $t - 1$. The spread of the sigma points is defined by $\kappa = \alpha^{2}(n + \gamma) - n$. The parameter $\alpha$ determines the spread, while $\gamma$ is a secondary scaling factor, typically set to one. The initial condition must be known $x_{0} \sim N(x_{0}, P_{0})$.

Then, the update process continues:
\begin{equation}
\begin{aligned}
    \bar{x}_{t|t-1} &= f \left( x_{t-1}, u_{t-1} \right) \\
    x_{t}^{-} &= \sum_{i = 0}^{2n} W_{i}^{m} \bar{x}_{i, t|t-1} \\
    P_{t}^{-} &= \sum_{i = 0}^{2n} W_{i}^{c} \left( \bar{x}_{i, t|t-1} - x_{t}^{-} \right) \left( \bar{x}_{i, t|t-1} - x_{t}^{-} \right)^{\top} + Q,
\end{aligned}
\end{equation}

where $\bar{x}_t$ represents the predicted mean, $\bar{P}_t$ represents the predicted covariance, {\em Q} is a covariance matrix that models the uncertainty in the system dynamics, and $W_i^m$ and $W_i^c$ are the weights for the mean and covariance, respectively, associated with the $i-{th}$ sigma point \cite{wan2000unscented}.

\[
W_0^m = \frac{\kappa}{n + \kappa}
\]
\[
W_0^c = \frac{\kappa}{n + \kappa} + (1 - \alpha^2 + \beta)
\]
\[
W_i^m = W_i^c = \frac{1}{2(n + \kappa)}, \quad i = 1, 2, \ldots, 2n
\]

where $\beta$ is a parameter that allows for incorporating prior knowledge about the distribution of the state vector $\mathbf{x}$. For Gaussian distributions, the optimal value for $\beta$ is 2.

\textbf{Measurement   Step:} In the measurement step, sigma points are calculated for the visual data and IMU data fusion, followed by measurement update:
\begin{equation}
    \bar{X}_{t} = 
    \begin{bmatrix}
        \bar{x}_{t} & \bar{x}_{t} \pm \sqrt{(n + \kappa) P_{t}}
    \end{bmatrix}
\end{equation}

\begin{equation}
\bar{Y}_{t} = g(\bar{X}_{t})
\end{equation}

\begin{equation}
\bar{y}_{t} = \sum_{i = 0}^{2n} W_{i}^{m} \bar{Y}_{i, t}
\end{equation}

\begin{equation}
P_{y_{t}} = \sum_{i = 0}^{2n} W_{i}^{c} (\bar{Y}_{i, t} - \bar{y}_{t}) \cdot (\bar{Y}_{i, t} - \bar{y}_{t})^{T} + R
\end{equation}

\begin{equation}
P_{x_{t}y_{t}} = \sum_{i = 0}^{2n} W_{i}^{c} (\bar{X}_{i, t} - \bar{x}_{t}) \cdot (\bar{Y}_{i, t} - \bar{y}_{t})^{T}
\end{equation}

\begin{equation}
K_{t} = P_{x_{t}y_{t}} \cdot \left( P_{y_{t}} \right)^{-1}
\end{equation}

\begin{equation}
v_{t} = y_{v, t} - \bar{y}_{t}
\end{equation}

\begin{equation}
x_{t} = x_{t}^{-}  + K_{t} \cdot v_{t}
\end{equation}

\begin{equation}
P_{t} = P_{t}^{-} - K_{t} \cdot P_{y_{t}} \cdot K_{t}^{T}
\end{equation}
where \(\bar{Y}_{t}\) represents sigma points projected through the measurement function \(g\), while \(\bar{y}_{t}\) represents the predicted measurement. The predicted measurement covariance \(P_{y_{t}}\) and the state-measurement cross-covariance \(P_{x_{t}y_{t}}\) are used to compute the Kalman gain \(K_{t}\), which updates the state vector \(x_t\) and its covariance \(P_t\).

The general pose estimation algorithm is shown in algorithm \ref{alg:UKF_IMU_Visual_Uncertainty}. This algorithm shows IMU and visual data fusion for pose estimation using a combination of DL and UKF. The DL models ViT for IMU features and MCNN for visual features extract features from the sensor data at each time step. The UKF predicts the state based on the previous state and then updates the state and covariance using fused IMU and visual features. Additionally, uncertainty estimates are incorporated into both the feature fusion and the state update processes, enhancing the robustness and reliability of pose estimation, particularly in challenging conditions where sensor data may be noisy or incomplete. The input to the model are IMU data \( \mathbf{a}_t, \boldsymbol{\omega}_t \) (accelerometer, gyroscope), visual data \( \mathbf{I}_t \) (image frame at time \( t \)), Initial state estimate \( \mathbf{x}_0 \), covariance \( \mathbf{P}_0 \), process noise covariance \( \mathbf{Q} \), measurement noise covariance \( \mathbf{R} \), and Uncertainty parameters \( \sigma_{\text{IMU}}, \sigma_{\text{Visual}} \). The model outputs final estimated pose \( \mathbf{x}_T \), covariance matrix \( \mathbf{P}_T \), and uncertainty estimation \( \sigma_T \).
%%%%%%%%%%%%%%%%%%%%%%%%%%%%%%%%%%%%%%%%%%%%%%%%%%%%%%%5
\begin{algorithm}[!ht]
\caption{IMU and Visual Data Fusion with Deep Learning for Pose Estimation and Uncertainty Estimation Using UKF}
\label{alg:UKF_IMU_Visual_Uncertainty}

\SetKw{KwInitialization}{Initialization}
\KwIn{ \( \mathbf{a}_t, \boldsymbol{\omega}_t \),  \( \mathbf{I}_t \), \( \mathbf{x}_0 \),  \( \mathbf{P}_0 \),  \( \mathbf{Q} \),  \( \mathbf{R} \), Uncertainty parameters (\( \sigma_{\text{IMU}}, \sigma_{\text{Visual}} \))}
\KwOut{Final estimated pose \( \mathbf{x}_T \), covariance matrix \( \mathbf{P}_T \), uncertainty estimation \( \sigma_T \)}

\KwInitialization{Initialize DL models: \( f_{\text{IMU}}(\mathbf{a}_t, \boldsymbol{\omega}_t) \),  \( f_{\text{visual}}(\mathbf{I}_t) \); \\ Initialize UKF parameters: \( \mathbf{x}_0 \), \( \mathbf{P}_0 \), and  UKF scaling factors \( \alpha, \kappa, \beta \);\\  Set uncertainty parameters \( \sigma_{\text{IMU}}, \sigma_{\text{Visual}} \)}

\For{each time step $t = 1, 2, \dots, T$}{
    \textbf{(a) DL-Based Feature Extraction:}\\
    Extract IMU features: \( \mathbf{z}_{\text{IMU}, t} = f_{\text{IMU}}(\mathbf{a}_t, \boldsymbol{\omega}_t) \); \\
    Extract visual features: \( \mathbf{z}_{\text{visual}, t} = f_{\text{visual}}(\mathbf{I}_t) \); \\
    Estimate uncertainty in features: \( \sigma_{\text{IMU}, t} = \text{UncertaintyEstimator}(\mathbf{z}_{\text{IMU}, t}) \); \\
    \( \sigma_{\text{visual}, t} = \text{UncertaintyEstimator}(\mathbf{z}_{\text{visual}, t}) \)
    
    \textbf{(b) Prediction Step (UKF):}\\
    Compute sigma points based on prior state \( \mathbf{x}_{t-1} \): 
    \( \bar{X}_{t-1} = \left[ \mathbf{x}_{t-1}, \mathbf{x}_{t-1} \pm \sqrt{n + \kappa} \, \mathbf{P}_{t-1} \right] \); \\
    Propagate sigma points through the process model: 
    \( \bar{x}_{t|t-1} = f_{\text{process}}(\bar{X}_{t-1}, \mathbf{u}_{t-1}) \); \\
    Predict state mean and covariance: \( \mathbf{x}_{t}^{-} = \sum_{i=0}^{2n} W_i^m \bar{x}_{i,t|t-1} \); \\
    \( \mathbf{P}_{t}^{-} = \sum_{i=0}^{2n} W_i^c \left( \bar{x}_{i,t|t-1} - \mathbf{x}_{t}^{-} \right) \left( \bar{x}_{i,t|t-1} - \mathbf{x}_{t}^{-} \right)^{\top} + \mathbf{Q} \)
    
    \textbf{(c) Measurement Update:}\\
    Fuse the IMU and visual features adaptively: 
    \( \mathbf{z}_t = \text{AdaptiveFusion}(\mathbf{z}_{\text{IMU}, t}, \mathbf{z}_{\text{visual}, t}, \sigma_{\text{IMU}, t}, \sigma_{\text{visual}, t}) \); \\
    Calculate sigma points based on predicted state \( \mathbf{x}_{t}^{-} \): 
    \( \bar{Y}_t = g(\bar{X}_t) \); \\
    Predict measurement mean and covariance: \( \mathbf{y}_t = \sum_{i=0}^{2n} W_i^m \bar{Y}_{i,t} \); \\
    \( \mathbf{P}_{y_t} = \sum_{i=0}^{2n} W_i^c \left( \bar{Y}_{i,t} - \mathbf{y}_t \right) \left( \bar{Y}_{i,t} - \mathbf{y}_t \right)^{\top} + \mathbf{R} \)
    
    \textbf{(d) Update State Estimate, Covariance, and Uncertainty:}\\
    Compute Kalman gain: \( \mathbf{K}_t = \mathbf{P}_{x_t y_t} \mathbf{P}_{y_t}^{-1} \); \\
    Update state estimate: \( \mathbf{x}_t = \mathbf{x}_{t}^{-} + \mathbf{K}_t \left( \mathbf{y}_t - \bar{y}_t \right) \); \\
    Update state covariance: \( \mathbf{P}_t = \mathbf{P}_{t}^{-} - \mathbf{K}_t \mathbf{P}_{y_t} \mathbf{K}_t^{\top} \); \\
    Update uncertainty estimate: \( \sigma_t = \text{UncertaintyEstimator}(\mathbf{x}_t, \mathbf{P}_t) \)
}
\textbf{Output:} Final estimated pose \( \mathbf{x}_T \), Covariance matrix \( \mathbf{P}_T \), Uncertainty estimate \( \sigma_T \)
\end{algorithm}

\subsection{Uncertainty-Aware Loss Function}
\label{uncertainty}
In addition to the proposed sensor fusion model that combines IMU and visual data with an adaptive DL approach, we incorporate uncertainty estimation. There are two main types of uncertainties: aleatoric and epistemic. Aleatoric uncertainty arises from inherent randomness in the data, such as sensor noise, blur, occlusion, etc.
Epistemic uncertainty occurs when there is insufficient knowledge during model training or prediction for a new test sample due to factors such as limited training samples, sub-optimal model architecture, or parameter learning issues. Handling such issues for robotics localization is essential because such problems affect efficient navigation due to sensor noise, missing data, etc. 
To achieve this, we introduce an uncertainty-aware loss function designed to enhance the effectiveness of feature learning by explicitly considering uncertainty. This loss function prioritizes minimizing uncertainty in the predictions, which is especially important in scenarios where sensor data may be unreliable, noisy, or incomplete.
By integrating uncertainty quantification directly into the model, we allow the system to focus on improving prediction accuracy and understanding the confidence level associated with each prediction. This two-fold emphasis results in a more robust and adaptive system, particularly under challenging or dynamic conditions where traditional approaches may struggle. Quantifying uncertainty allows the system to more accurately assess the reliability of the data, empowering it to adapt its actions accordingly.

For example, the model can recognize increased uncertainty in its predictions when the IMU or visual sensor data are compromised by environmental noise, sensor degradation, or occlusion. The uncertainty-aware loss function then encourages the model to prioritize reducing this uncertainty, ensuring that the system can still function reasonably even under degraded conditions.
This capability is especially beneficial for autonomous navigation in environments where sensor reliability may fluctuate, such as tunnels or areas with limited visibility. The adaptive learning mechanism with the UKF continuously updates the pose estimation and fuses sensor data to correct for drift or errors. By factoring in uncertainty, the UKF can weigh sensor inputs more appropriately, favoring data with lower associated uncertainty and thus improving the final pose estimate.
Moreover, the integration of uncertainty quantification enhances the overall robustness of the navigation system. In critical decision-making scenarios, where the vehicle must navigate complex routes, or  avoid obstacles, measuring and accounting for uncertainty provides a significant advantage. The system improves detection accuracy and assesses the confidence level of the pose estimation, making the navigation process more transparent and trustworthy. The model is trained and tested by adding noises for robust navigation to handle such cases.

The proposed uncertainty-aware loss function is designed to optimize both the pose prediction accuracy and the associated uncertainty. It combines the traditional \textit{pose loss} (translation and rotation) with a weighting mechanism based on predicted uncertainty. This approach encourages the model to minimize prediction error while also penalizing overconfidence in uncertain predictions.

\subsubsection{Pose Loss}

The pose loss is computed as the Mean Squared Error (MSE) between the ground truth and predicted values for both translation and rotation:

\begin{equation}
\mathcal{L}_{\text{pose, trans}} = \frac{1}{N} \sum_{i=1}^{N} \| \mathbf{p}_{\text{truth},i} - \mathbf{p}_{\text{pred},i} \|^2
\end{equation}

\begin{equation}
\mathcal{L}_{\text{pose, rot}} = \frac{1}{N} \sum_{i=1}^{N} \| \mathbf{e}_{\text{truth},i} - \mathbf{e}_{\text{pred},i} \|^2
\end{equation}

The combined pose loss is:

\begin{equation}
\mathcal{L}_{\text{pose}} = \mathcal{L}_{\text{pose, trans}} + \mathcal{L}_{\text{pose, rot}}
\end{equation}
Where:
\begin{itemize}
    \item \( \mathbf{p}_{\text{truth}} \in \mathbb{R}^3 \) be the ground truth translation vector.
    \item \( \mathbf{e}_{\text{truth}} \in \mathbb{R}^3 \) be the ground truth rotation vector (Euler angles).
    \item \( \mathbf{p}_{\text{pred}} \in \mathbb{R}^3 \) be the predicted translation vector.
    \item \( \mathbf{e}_{\text{pred}} \in \mathbb{R}^3 \) be the predicted rotation vector.
    \item {\em N} is the number of samples.
\end{itemize}
\subsubsection{Uncertainty-Aware Loss}

The uncertainty-aware loss incorporates the predicted uncertainty \( \sigma_{\text{trans}} \) and \( \sigma_{\text{rot}} \), where higher uncertainty reduces the weight of the pose error, while lower uncertainty increases it. This is formulated as:

\begin{equation}
\begin{aligned}
\mathcal{L}_{\text{uncertainty}} = \exp(-\sigma_{\text{trans}}) \cdot \mathcal{L}_{\text{pose, trans}} + &\\
\exp(-\sigma_{\text{rot}}) \cdot \mathcal{L}_{\text{pose, rot}} + & \\
(\sigma_{\text{trans}} + \sigma_{\text{rot}})
\end{aligned}
\end{equation}

Where:
\begin{itemize}
    \item \( \exp(-\sigma_{\text{trans}}) \) and \( \exp(-\sigma_{\text{rot}}) \) adjust the weight of the loss based on the uncertainty.
    \item \( \sigma_{\text{trans}} \) and \( \sigma_{\text{rot}} \) act as regularizers, penalizing predictions with excessive uncertainty.
\end{itemize}

\subsubsection{Total Loss}

The total loss is computed as the average of the uncertainty-aware loss over all the training samples:

\begin{align}
\mathcal{L}_{\text{total}} = \frac{1}{N} \sum_{i=1}^{N} &\left[ \exp(-\sigma_{\text{trans},i}) \cdot \mathcal{L}_{\text{pose, trans},i} \right. \nonumber \\
&\left. + \exp(-\sigma_{\text{rot},i}) \cdot \mathcal{L}_{\text{pose, rot},i} \right. \nonumber \\
&\left. + (\sigma_{\text{trans},i} + \sigma_{\text{rot},i}) \right]
\end{align}

This loss function encourages the model to optimize both the pose accuracy and the reliability of its uncertainty prediction, making it more robust to sensor noise, missing data, and environmental challenges.

By incorporating uncertainty estimation into the sensor fusion process and the loss function, the proposed model adapts more effectively to challenging conditions, such as occlusion or environments with limited visibility, as demonstrated in the experiment section \ref{sec:experiments}. This approach has numerous advantages. First, it allows for dynamic adaptation, where the model dynamically weighs sensor inputs, giving preference to data with lower uncertainty and improving pose estimation accuracy. Second, it enhances robustness to noisy data, as the uncertainty-aware loss enables the model to prioritize reducing uncertainty, allowing it to function effectively even in noisy or degraded conditions. Lastly, integrating uncertainty leads to improved decision-making, providing more accurate and reliable navigation, which enhances the system's ability to make critical decisions in complex environments.

\subsection{Evaluation Metrics}
The commonly used evaluation metrics for VO and VIO include the absolute trajectory error (ATE) and relative pose error (RPE) \cite{zhang2018tutorial, burgard2009comparison, sturm2012benchmark}. 

The ATE represents the root mean square error (RMSE) between the estimated pose $\hat{x}_i$, and the ground truth pose $x_i$. It quantifies the absolute deviation of the estimated pose from the true pose, considering the entire trajectory.
\begin{equation}
ATE_{trans} = \left( \frac{1}{N} \sum_{i=0}^{N-1} \| \Delta t_i \|^2 \right)^{\frac{1}{2}}
\label{eq:eqn21}
\end{equation}
\begin{equation}
ATE_{rot} = \left( \frac{1}{N} \sum_{i=0}^{N-1} \| \Delta R_i \|^2 \right)^{\frac{1}{2}}
\label{eq:eqn20}
\end{equation}

On the other hand, the RPE is the RMSE of the relative pose difference between two consecutive frames $i$ and $i+1$:
% RPE Translation
\begin{equation}
   \text{RPE}_{\text{trans}} = \left( \frac{1}{N-1} \sum_{i=0}^{N-2} \|\Delta \mathbf{t}_{i, i+1} \|^2 \right)^{\frac{1}{2}} 
\end{equation}

% RPE Rotation
\begin{equation}
   \text{RPE}_{\text{rot}} = \left( \frac{1}{N-1} \sum_{i=0}^{N-2} \|\Delta \mathbf{R}_{i, i+1} \|^2 \right)^{\frac{1}{2}} 
\end{equation}

 Where, \( N \) represents the total number of frames, while \( \Delta R_i \) and \( \Delta t_i \) denote the rotational and translational errors between the estimated pose  and the ground truth pose, respectively. 

\section{Experiments}
\label{sec:experiments}
The experiment uses the KITTI visual-inertial dataset \cite{Geiger2013IJRR}. 
The dataset comprises various driving conditions, providing a dynamic and challenging benchmark for improving the accuracy and robustness of autonomous driving.
The dataset contains 22 video sequences, with sequences 00–10 having ground truth trajectories and sequences 11–22 consisting of raw data without ground truth, used for evaluation. The proposed model was trained on sequences 00, 01, 02, 05, 06, 08, and 09 and tested on sequences 04, 07, and 10. Sequence 03 was unavailable during training and, therefore, was omitted. All experiments were conducted using NVidia A100/80G GPUs running on a Linux OS.

\begin{table*}[!ht]
\caption{ATE and RPE results on the sequences 04, 07, and 10. 'trans' and 'rot' refer to translation and rotation. * shows the ATE results irrespective of translation or rotation.}
\label{tab:table1}
\resizebox{\textwidth}{!}{%
\begin{tabular}{|l|ll|ll|cl|ll|ll|ll|}
\hline
\multirow{2}{*}{Method} & \multicolumn{4}{c|}{04}                   & \multicolumn{4}{c|}{07}                   & \multicolumn{4}{c|}{10}                   \\ \cline{2-13} 
                        & ATE-trans & ATE-rot & RPE-trans & RPE-rot & ATE-trans & ATE-rot & RPE-trans & RPE-rot & ATE-trans & ATE-rot & RPE-trans & RPE-rot \\ \hline
VISO2 \cite{geiger2011stereoscan}&   5.5588*      &    -       &       0.0282  &    0.0202       &    18.1500*     & -  &  0.1182      &  0.0410                  & 47.8139* & -&0.2241&0.0305      \\
ORB-SLAM2 \cite{mur2017orb}&     1.4075*      &  -       &   0.0136        &     0.0019    &    16.2925*       &    -     &         0.0924  &0.0030         &    5.4202*       &      -   &      0.0336     & 0.0028        \\
sfmLearner \cite{zhou2017unsupervised} &      2.9668*     &   -      &    0.0462       &    0.0197     &  21.4758*         &    -     &   0.1082        &   0.0452      &       20.2817    &    -     &      0.1068     & 0.0447        \\
 Depth-VO-Feat \cite{zhan2018unsupervised} &    1.5141*       &  -       &       0.0253    & 0.0189       &    15.2080*       &    -     &    0.1858       &  0.0304       &    12.9822*       &  -       &       0.22754    &  0.027       \\
 GD-VIO \cite{yusefi2023generalizable}  &    1.4919*       &   -     &         0.0272  &  0.0176       &  16.7930*         &    -     &     0.3801      &0.0743         &     24.2930*      &   -      &      0.4173     &  0.0562       \\
    Proposed-Adaptive  &  0.8064         &   0.5663      &  0.0046         &  0.0013       &10.3970           &  7.9250       &    0.0084       &   0.0204      &  11.8292         &  10.2275       &   0.0160        &   0.0020      \\ \hline
\end{tabular}%
}
\end{table*}
%%%%%%%%%%%%%%%%%%%%%%%%%%%%%%%%%%%%%%%%%%%%%%%%%%%%
The proposed model is efficient and suitable for realtime applications, especially on edge devices. The computational complexity is only 1.587 billion floating-point operations per second (GFLOPs), and the model processes a single frame in 6.432 milliseconds (155 frames per second), indicating it uses minimal power while providing fast processing speeds. Its parameter counts are 6.983 million, and it uses only 173.755 MB of memory, making it easy to deploy on devices with limited resources, like mobile GPUs and embedded systems. To evaluate the robustness of the proposed model, we conducted tests under various degradation scenarios that simulate real-world challenges faced in autonomous navigation. Each type of degradation corresponds to potential environmental or sensor-related issues that autonomous systems might encounter. 
The dataset was subjected to various degradation techniques to assess the model's robustness. The baseline data, called "Standard Condition," was used without degradation. To simulate partial visibility loss, mimicking scenarios where obstacles obscure the camera view, we introduced "Occlusion" by overlaying a 128×128 pixel mask on the sample images at random locations for each instance. To replicate motion blur or the effect of low-quality camera captures in high-speed situations, we apply "Image Blurring" using Gaussian blur with a $\sigma$=20 pixels, along with additional salt-and-pepper noise. These effects simulate the motion blur and noise when the camera or lighting conditions change significantly \cite{couzinie2013learning, chen2019selective}. We introduced "Frame Omission" by randomly removing 10\% of the input images to simulate temporary visual sensor failures, reflecting real-world scenarios where data might be lost or corrupted. This effect can also mimic conditions in areas with very poor illumination, such as tunnels. 
Similarly, IMU data was altered by adding random noise and bias to reflect sensor inaccuracies and drift, categorized as "IMU Noise."
Additionally, "IMU Missing" involved randomly omitting IMU data points to mimic intermittent IMU sensor outages. 
Lastly, the most challenging scenario, "All Degradation," combined all these degradation types to evaluate the model under extreme conditions.

Table \ref{tab:table1} compares the ATE and RPE metrics for various VO and VIO methods across sequences 04, 07, and 10, focusing on translation and rotation. The proposed adaptive method consistently shows the best performance across most metrics, particularly in minimizing ATE and RPE, making it the most accurate and reliable of the evaluated methods.
\begin{table*}[!ht]
\caption{Effect of Sensor Data Degradation on VIO Fusion}
\label{tab:table2}
\resizebox{\textwidth}{!}{%
\begin{tabular}{|l|ll|ll|cl|ll|ll|ll|}
\hline
\multirow{2}{*}{Degradation} & \multicolumn{4}{c|}{04}                   & \multicolumn{4}{c|}{07}                   & \multicolumn{4}{c|}{10}                   \\ \cline{2-13} 
                        & ATE-trans & ATE-rot & RPE-trans & RPE-rot & ATE-trans & ATE-rot & RPE-trans & RPE-rot & ATE-trans & ATE-rot & RPE-trans & RPE-rot \\ \hline
Standard Condition &0.8064         &   0.5663      &  0.0046         &  0.0013       &10.3970           &  7.9250       &    0.0084       &   0.0204      &  11.8292         &  10.2275       &   0.0160        &   0.0020 \\
Occlusion&    0.8227       &     0.6403    & 0.0062          & 0.0020        & 10.3990          &8.0200         &0.0088           &0.0206         &11.8447           & 10.5020        &0.0204           &0.0026         \\
Image Blurring &  0.8228         &0.6411         &         0.0062  &0.0022         &10.4405           &    8.0332     & 0.0091          & 0.0207        &11.8449           &10.5021         & 0.0205          & 0.0026        \\
Frame Omission &   1.0220        & 0.8947        &0.0977           &0.0094         &10.8804           & 8.0766        &0.0255           & 0.0860        &12.0460           & 10.8705        & 0.0623          &0.0716         \\
IMU Noise &   0.8230        &0.6414         &0.0066           &0.0023         &10.4407           & 8.0336        &0.0094           &0.0209         &11.8452           &10.5022         &0.0209           &0.0028         \\
IMU Missing    &   0.8235        & 0.6415        & 0.0068          & 0.0023        &10.4407           & 8.0337        &0.0102           &0.0220         & 11.8454          &10.5023         & 0.2011          & 0.0028        \\
All Degradation &   0.8828        &0.7016         &0.0347           &0.0056         &  10.5526         &8.0442         &0.01462           &0.0354         & 11.9224          &10.5582         &0.0304           &0.0184         \\ \hline
\end{tabular}%
}
\end{table*}
\begin{figure*}
    \centering
    \includegraphics[width=0.99\linewidth]{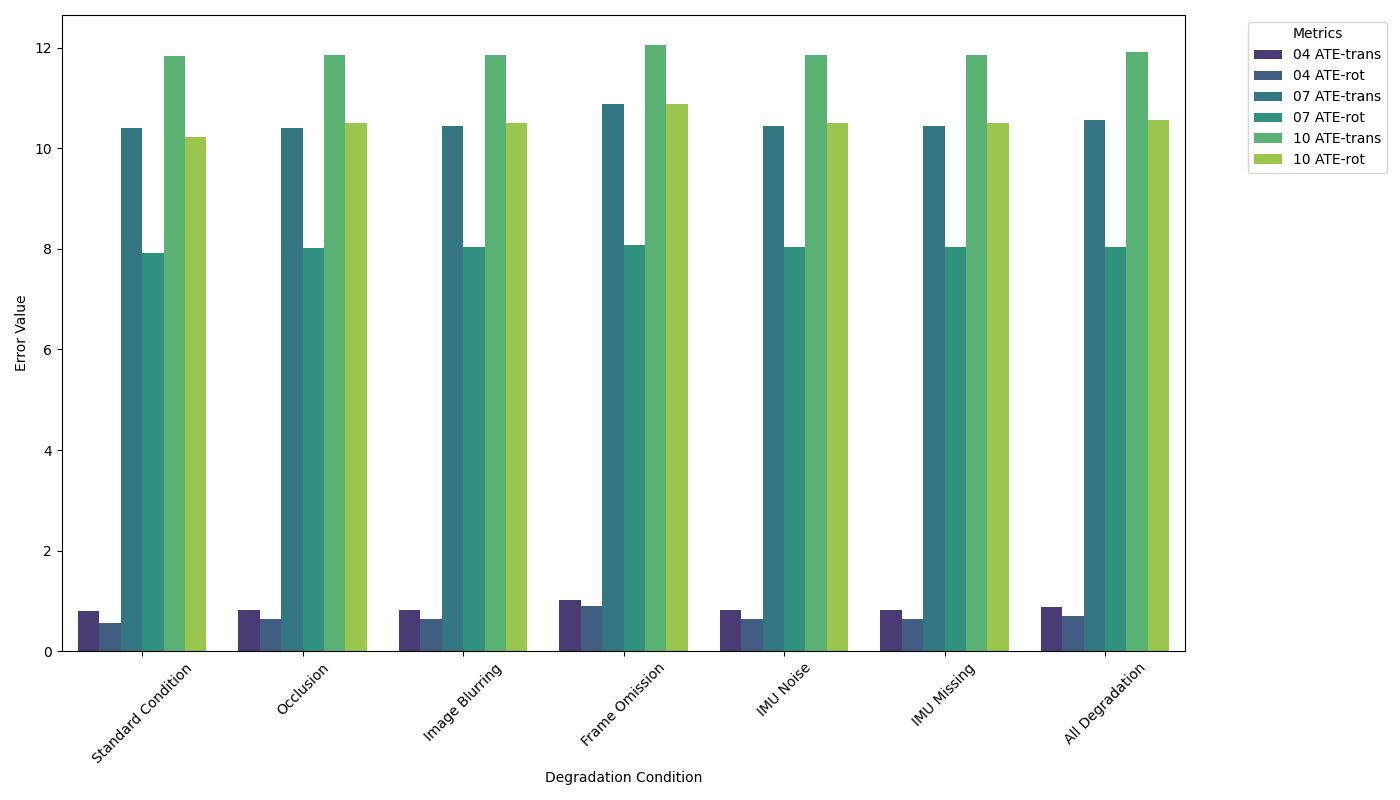}
    \caption{ATE and RPE  comparison across sequences 04, 07, and 10. The {\em x}-axis represents different sensor data degradations on VIO fusion performance to simulate real-world scenarios. The {\em y}-axis indicates the resulting rotation and translation errors associated with these data degradations.}
    \label{fig:fig4}
\end{figure*}
Table \ref{tab:table2} and Fig. \ref{fig:fig4} demonstrates how various sensor data degradations can impact the performance of VIO fusion, reflecting real-world scenarios that might occur in actual systems. Such degradations, including occlusion, image blurring, frame omission, IMU noise, and missing IMU data, mimic challenges a real VIO system may encounter, such as environmental interference or sensor failures. The Standard Condition, which represents the ideal scenario without degradation, consistently delivers the best performance across all sequences, achieving the lowest errors in ATE and RPE for translation and rotation.
When degradations are introduced, performance declines significantly, highlighting the system's sensitivity to specific conditions. For example, Frame Omission significantly impacts performance, leading to the highest error rates in sequences 04 and 07, where ATE-trans rises to 1.0220 and 10.8804, and RPE-trans increases to 0.0977 and 0.0255, respectively. Similarly, the All Degradation condition, which combines multiple degradation factors, leads to the highest errors in sequence 10, with an ATE-trans of 11.9224 and RPE-trans of 0.0304. Other degradations, such as Image Blurring, IMU Noise, and IMU Missing, have a more moderate impact, but they still degrade the system's accuracy compared to the Standard Condition.

%%%%%%%%%%%%%%%%%%%%%%%%%%%%%%%%%%%%%%%%%%
\begin{table*}[!ht]
\caption{Contribution of Each Module to Overall Model Performance }
\label{tab:table3}
\resizebox{\textwidth}{!}{%
\begin{tabular}{|l|ll|ll|cl|ll|ll|ll|}
\hline
\multirow{2}{*}{Method} & \multicolumn{4}{c|}{04}                   & \multicolumn{4}{c|}{07}                   & \multicolumn{4}{c|}{10}                   \\ \cline{2-13} 
                        & ATE-trans & ATE-rot & RPE-trans & RPE-rot & ATE-trans & ATE-rot & RPE-trans & RPE-rot & ATE-trans & ATE-rot & RPE-trans & RPE-rot \\ \hline

Standard Fusion &0.8116&0.6030&0.0048&0.0015&10.4407&8.0020&0.0088&0.0207&11.8424&10.2355&0.0192&0.0042\\
Adaptive Fusion&   0.8092        &    0.5677     &   0.0051        &   0.0014      &      10.3982     &   7.9266      &     0.0088      &    0.0205     &   11.8302        &   10.2284      &  0.0169         &   0.0025      \\
Uncertainty-Aware Loss &   0.8068        & 0.5667        &   0.0048        &   0.0014      &   10.3990        &   7.9258      &   0.0086        &   0.0204      &       11.8297    &   10.2278      &    0.0165       &   0.0022      \\
Vision Only &0.8117&0.6032&0.0053&0.0015&10.4410&7.9986&0.0094&0.0207&11.8429&10.2365&0.0199&0.0038\\
All  &  0.8064         &   0.5663      &  0.0046         &  0.0013       &10.3970           &  7.9250       &    0.0084       &   0.0204      &  11.8292         &  10.2275       &   0.0160        &   0.0020   \\
 \hline
\end{tabular}%
}
\end{table*}
 Table \ref{tab:table3} analyzes how various individual modules contribute to the model's overall performance on the KITTI VIO dataset across sequences 04, 07, and 10. 
In sequence 04, the adaptive fusion method reduces the translational error (ATE-trans) to 0.8092, an outperforming standard fusion with an ATE-trans of 0.8116. The uncertainty-aware loss method yields similar outcomes with an ATE-trans of 0.8068 and achieves the lowest RPE-trans (0.0048) and RPE-rot (0.0014), indicating its effectiveness in managing fine-grained relative pose errors. Conversely, the vision-only method shows higher errors, with RPE-trans at 0.0053 and RPE-rot at 0.0015, highlighting its limitations compared to the fusion methods.

For sequence 07, both adaptive fusion and uncertainty-aware loss show strong performance. Adaptive fusion reduces ATE-rot (7.9266) and RPE-rot (0.0205), while the uncertainty-aware loss method slightly enhances these results (ATE-rot: 7.9258, RPE-rot: 0.0204). The vision-only method reduces performance, especially in the translational metrics (ATE-trans: 10.4410, RPE-trans: 0.0094), emphasizing the advantages of multi-sensor fusion.
In sequence 10, adaptive fusion and uncertainty-aware loss maintain their strong performance. Adaptive fusion results in an ATE-rot of 10.2284 and an RPE-rot of 0.0025, while the uncertainty-aware loss method reaches slightly better results with an RPE-rot of 0.0022. Standard fusion and vision-only continue to show higher errors across all metrics, reinforcing the necessity for adaptive and uncertainty-aware approaches to enhance overall performance.

Ultimately, when all modules are integrated into the "All" configuration, the system shows the lowest error values across nearly all metrics and sequences. This analysis underscores the importance of designing robust VIO systems that can handle real-world degradation scenarios. By understanding how these factors affect performance, strategies can be developed to mitigate their impact and enhance the reliability of VIO systems in practical applications.
\section{Conclusion}
\label{sec:conclusion}
This work presents a lightweight, efficient, and optimized hybrid deep learning and UKF  model to improve pose estimation for autonomous driving. To ensure a thorough evaluation, we introduced various sensor data degradation scenarios that simulate real-world challenges, such as image occlusion, image blurring, IMU noise, frame omission, and missing IMU data. These degradations reflect potential environmental and sensor failures in autonomous systems, such as temporary loss of visibility, motion blur, and inertial sensor inaccuracies.
The results demonstrate that the proposed adaptive fusion and uncertainty-aware loss approaches consistently outperform baseline methods, particularly in minimizing ATE and RPE for translation and rotation. Under ideal conditions, the system achieves the lowest error rates across all metrics, providing an upper bound on accuracy. However, the performance degrades significantly under challenging conditions, such as Frame Omission, Image Blurring, and  IMU Noise. 
The ablation study highlights the contributions of individual modules to overall performance. The adaptive fusion and uncertainty-aware loss methods consistently achieve lower translational and rotational errors across sequences. For instance, uncertainty-aware loss achieves the lowest RPE-trans (0.0048) and RPE-rot (0.0014) in sequence 04, while adaptive fusion reduces ATE-trans to 0.8092. Both methods maintain strong performance in sequences 07 and 10, validating their robustness in handling complex and noisy environments. In contrast, the vision-only approach shows higher errors, reinforcing the importance of multi-sensor fusion for achieving reliable pose estimation. 

The results emphasize the critical need for robust VIO systems capable of handling real-world challenges. By systematically evaluating the impact of various degradations and analyzing the contributions of individual modules, this work provides key insights into mitigating the effects of sensor failures and environmental interference. The proposed system’s resilience and superior performance make it a promising solution for autonomous navigation in dynamic and real-world environments. Future work can focus on further enhancing system robustness through advanced sensor fusion techniques, uncertainty quantification, and strategies to address long-term sensor drift.

%\newpage
% \vspace{-1pt}
% \begin{IEEEbiography}
% [{\includegraphics[width=1in,height=1.25in,clip,keepaspectratio]{author1.jpg}}]
% {Simegnew Yihunie Alaba} (M'19--SM'26) received a Ph.D. degree in Electrical and Computer Engineering from Mississippi State University in 2024. His academic focus encompasses machine learning, deep learning, and computer vision for autonomous driving and robotics. 
% \end{IEEEbiography}
% \vspace{-2pt}
% \begin{IEEEbiography}[{\includegraphics[width=1in,height=1.25in,clip,keepaspectratio]{author2.jpg}}]
% {Yuichi Motai}  (S'00--M'03--SM'12) received the B.Eng. degree in instrumentation engineering from Keio University, Tokyo, Japan, in 1991, the M.Eng. degree in applied systems science from Kyoto University, Kyoto, Japan, in 1993, and the Ph.D. degree in electrical and computer engineering from Purdue University, West Lafayette, IN, U.S.A., in 2002. He is currently an Associate Professor of Electrical and Computer Engineering at Virginia Commonwealth University, Richmond, VA, U.S.A. His research interests include the broad area of sensory intelligence; particularly in data analytics, pattern recognition, computer vision, and sensory-based robotics.
% \end{IEEEbiography}

% \vfill

\end{document}